\documentclass[letterpaper]{article} 
\usepackage[preprint]{aaai2027}
\usepackage[hyphens]{url}  
\usepackage{graphicx} 
\usepackage{natbib}  
\usepackage{caption} 
\usepackage{algorithm}
\usepackage{algorithmic}

\usepackage{newfloat}
\usepackage{listings}
\DeclareCaptionStyle{ruled}{labelfont=normalfont,labelsep=colon,strut=off} 
\floatstyle{ruled}
\newfloat{listing}{tb}{lst}{}
\floatname{listing}{Listing}

\usepackage{booktabs}

\usepackage{amsmath}
\usepackage{amssymb}
\usepackage{array}
\usepackage[table]{xcolor}
\usepackage{tabularx}
\usepackage{multirow}

\newcolumntype{L}[1]{>{\raggedright\arraybackslash}m{#1}}
\newcolumntype{C}[1]{>{\centering\arraybackslash}m{#1}}
\newcolumntype{Y}[1]{>{\hsize=#1\hsize\linewidth=\hsize\centering\arraybackslash}X}

\title{PEAM: On-device Communication Agent for Personalized, Efficient and Autonomous Mobile Network Configuration}
\author{
    Haowen Wan\textsuperscript{\rm 1,2},
    Shangzhuo Xie\textsuperscript{\rm 2},
    Syue-Ju Syue\textsuperscript{\rm 3},
    Xianyue Wu\textsuperscript{\rm 3},
    Runhua Li\textsuperscript{\rm 3},
    Qianqian Yang\textsuperscript{\rm 2}
}
\affiliations{
    \textsuperscript{\rm 1}Polytechnic Institute, Zhejiang University\\
    \textsuperscript{\rm 2}College of Information Science and Electronic Engineering, Zhejiang University\\
    \textsuperscript{\rm 3}OPPO\\
    
}

\begin{document}

\maketitle
\pagestyle{plain}
\thispagestyle{plain}

\begingroup
\renewcommand{\thefootnote}{}
\footnotetext{%
    \footnotesize
    Qianqian Yang is responsible for
    correspondence regarding this paper. Email: qianqianyang20@zju.edu.cn.
}
\endgroup

\begin{abstract}
The growing deployment of compact large language models on smartphones is enabling autonomous on-device decision-making, such as emerging mobile GUI agents. However, many mobile decisions, such as network configuration when users switch apps or connectivity conditions change, must satisfy stringent latency constraints to preserve user experience. In this paper, we propose PEAM, an on-device LLM-based communication agent that autonomously generates personalized network-configuration rules executable within milliseconds. Specifically, PEAM first fine-tune the on-device LLM with a dataset constructed from manually annotated real-user data to integrate the communication knowledge into the local model. PEAM then operates in two phases. In the rule-generation phase, which is performed when the device is idle, such as overnight, PEAM analyzes historical network-configuration records in batches and reflects across batches to consolidate consistent evidence, reduce local bias, and derive robust configuration rules. To enable more accurate and stable rule generation over time, PEAM further constructs a user profile from high-confidence rules to capture the user’s network preferences and incorporates this profile into each rule-generation cycle. In the rule-execution phase, PEAM is invoked when the communication context changes, such as when users switch apps or connectivity conditions change, and searches the rule library to identify and execute the appropriate rule. Experiments demonstrate that PEAM, based on a fine-tuned 4B model, achieves 99.13\% correct network configuration decisions, while 92.50\% of the rules generated by PEAM are valid, outperforming the traditional expert-rule based method by 4.41\% and 84.50\%, respectively.

\end{abstract}

\section{Introduction}
Recent advances in large language models (LLMs) have significantly improved their reasoning, planning, and autonomous decision-making capabilities. Meanwhile, the growing deployment of compact models on mobile and edge devices has made LLM-based intelligence increasingly accessible in resource-constrained environments \cite{mnn, confidant, mobile_p_agent}, as demonstrated by emerging mobile GUI agents\cite{autodroid, agentCPM, ferret}. These trends open up new opportunities for using on-device LLMs to support complex decision-making tasks in mobile systems. One representative task is mobile network configuration, which dynamically selects and adjusts communication resources and settings, such as access networks, WiFi access points, and SIM cards, according to application requirements, network conditions, device states, and user contexts. Unlike many offline decision-making tasks, mobile network configuration directly affects user experience and often lies on a latency-critical execution path. A practical solution should therefore be \textbf{personalized}, because different users may prefer different configurations under similar communication conditions; \textbf{efficient}, because configuration decisions must often be completed within stringent latency constraints; and \textbf{autonomous}, because the device should learn and apply user preferences without repeated manual intervention.

\begin{figure}[!t]
    \centering
    \includegraphics[width=1.0\linewidth]{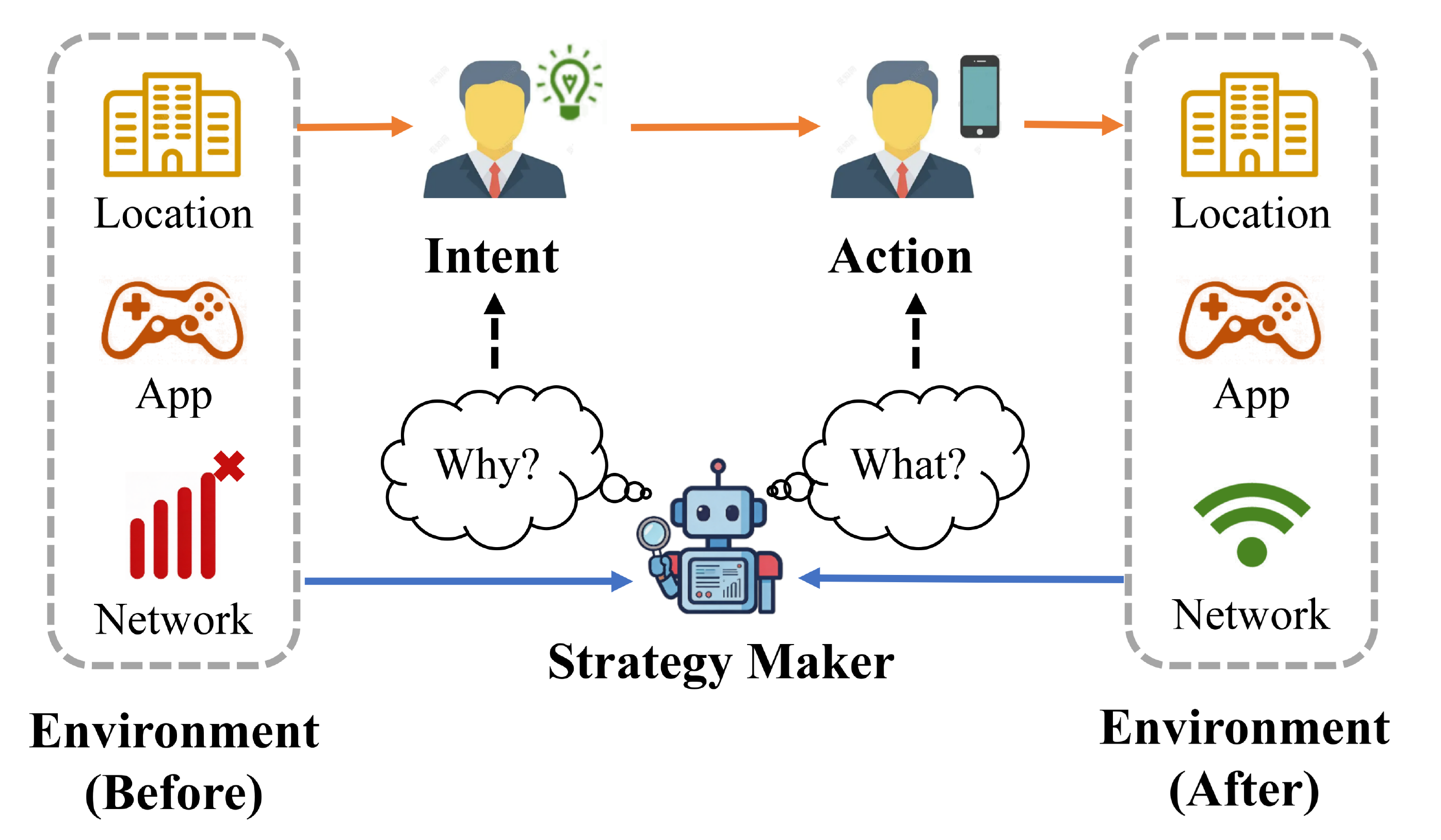}
    \caption{Modeling personalized mobile network configuration from historical decision records. PEAM infers why a user prefers a particular configuration and determines what configuration action should be executed.}
    \label{fig:1}
\end{figure}

Existing network-configuration methods mainly fall into three categories: expert-rule (ER)-based, supervised-learning (SL)-based, and reinforcement-learning (RL)-based methods. ER-based methods remain widely used in practical systems. They encode communication knowledge into hand-crafted constraints over metrics such as signal strength, uplink/downlink latency, and quality of service (QoS), and then map such conditions to switching actions\cite{er_fuzzy, er_ahp, er_autoplex}. Although ER-based methods are efficient at runtime, their strategies are primarily determined by expert-designed heuristics, making them difficult to personalize according to individual user behavior. To improve adaptability, recent studies have explored SL-based and RL-based methods. SL-based methods learn correlations between communication observations and user actions from labeled data\cite{dl_xgboost, dl_dnn, dl_cnn}, whereas RL-based methods formulate network switching as a sequential decision process and optimize policies through reward-driven interaction\cite{rl_ql, rl_dqn, rl_2ddqn, rl_abuv, rl_adbits, rl_udn}. However, in these methods, user preferences are still implicitly encoded in hand-crafted rules, training labels, reward functions, or model parameters. As a result, they are difficult to interpret, update, and continually adapt as user behavior evolves.


More recently, researchers have begun to apply LLMs to network-configuration tasks. Existing studies have explored LLM-based solutions for WiFi roaming\cite{roaming_agent, roaming_agent2} and cellular handover\cite{cellular_agent, cellular_agent2}, demonstrating the potential of LLMs to interpret heterogeneous communication observations and generate corresponding decisions. However, these approaches largely employ LLMs as reactive decision engines that directly map current observations to configuration actions. They do not systematically distill long-term user behavior into persistent, interpretable configuration rules, and they still rely on LLM inference during the decision-making path. Consequently, existing LLM-based methods struggle to simultaneously achieve personalization, efficiency, and autonomy for on-device mobile network configuration.


To address these limitations, we propose PEAM, an on-device LLM-based communication agent for personalized, efficient, and autonomous mobile network configuration. The key idea of PEAM is to decouple complex LLM reasoning from latency-critical execution. Instead of invoking the LLM for every online configuration decision, PEAM performs LLM-based reasoning offline when the device is idle, distills historical user behavior into persistent condition--action rules, and executes these rules online through lightweight retrieval.  As illustrated in Figure~\ref{fig:1}, PEAM models personalized network configuration as two coupled problems: understanding why a user prefers a particular configuration from historical records and determining what configuration action should be executed under the current communication context.

This design introduces two key challenges on resource-constrained devices. First, \textbf{how can a compact on-device LLM reliably infer user intents and generate configuration rules from long and heterogeneous historical records?} General-purpose LLMs lack communication-specific knowledge. Moreover, on-device compact models usually have limited context capacity. PEAM addresses this challenge by constructing a task-specific dataset from manually annotated real-user network-switching records and performing supervised fine-tuning to integrate communication knowledge into the local model. PEAM then partitions long histories into coherent batches, analyzes each batch independently, and performs cross-batch reflection over the intermediate outputs. This divide-and-consolidate process keeps each inference within the model’s context window while integrating distributed evidence, reducing locally biased conclusions, and producing more reliable configuration rules.

Second, \textbf{how can we continually capture evolving user preferences and improve the accuracy and stability of rule generation over time?} Static rules cannot adequately represent preferences that change with user behavior. PEAM therefore updates rule confidence according to whether the user accepts or rejects the configuration proposed by the retrieved rule, and discards rules whose confidence falls below a predefined threshold. Beyond maintaining individual rules, PEAM constructs a semantic user profile from high-confidence rules. This profile summarizes both scenario-specific preferences, such as preferred configurations under particular applications, times, locations, and network conditions, and broader behavioral tendencies, such as sensitivity to data usage and preference for connection quality. Rather than directly inheriting the previous profile, PEAM periodically regenerates the profile from the latest reliable rules and incorporates it into the next rule-generation cycle, providing an up-to-date preference context for more accurate and stable personalization.


The main contributions of this paper are summarized as follows:
\begin{itemize}
    \item We propose PEAM, an on-device communication agent that distills offline LLM reasoning into persistent condition--action rules and executes them through lightweight online retrieval. This design decouples complex deliberation from latency-critical execution, providing a practical architecture for LLM-based decision-making on resource-constrained devices.
    \item We design a divide-and-consolidate reasoning mechanism that decomposes long, heterogeneous interaction histories into coherent batches and consolidates distributed evidence across batches, enabling compact on-device LLMs to generate personalized rules beyond a single context window.
    \item We develop a rule-based hybrid memory that combines executable rule memory with semantic preference memory. By reconstructing user profiles from high-confidence rules and feeding them into subsequent reasoning cycles, PEAM supports interpretable and behavior-based personalization over time.
    \item We construct benchmarks based on real-user and simulated-user records for comprehensive evaluation. Extensive results, together with a smartphone deployment of a quantized 4B model, demonstrate PEAM’s effectiveness and practical on-device feasibility.
\end{itemize}

\begin{figure*}
    \centering
    \includegraphics[width=1\linewidth]{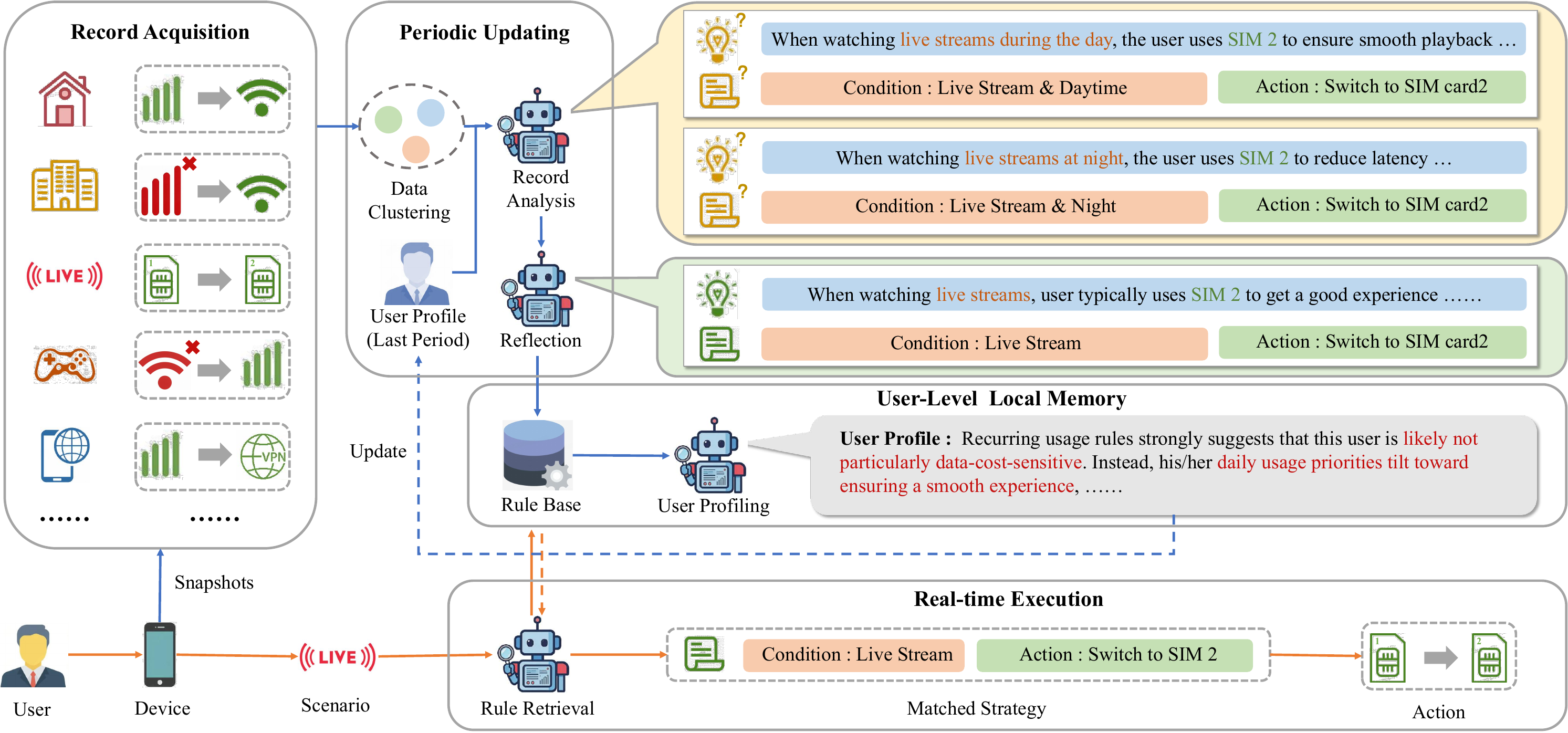}
    \caption{Overview of PEAM. The framework contains four stages: (1) Record Acquisition collects user-initiated configuration records; (2) Periodic Update performs clustering, batch-wise record analysis, and cross-batch reflection to generate personalized rules; (3) User-Level Local Memory maintains the rule base and user profile; and (4) Real-Time Execution retrieves and executes matched rules without invoking the LLM.}
    \label{fig:2}
\end{figure*}

\section{Related Work}

\noindent\textbf{On-device LLM-based Agent}\quad On-device LLM agents have recently attracted increasing attention for enabling autonomous perception, reasoning, and interaction directly on mobile devices. Advances in mobile inference and collaborative model customization provide important technical foundations for such agents. For example, MNN \cite{mnn} supports efficient neural network inference across heterogeneous mobile hardware, while Confidant \cite{confidant} enables collaborative customization of Transformer-based LLMs across multiple mobile devices.At the application level, mobile GUI automation has become a representative direction for on-device agents. AutoDroid \cite{autodroid} formulates GUI automation as code generation, allowing an on-device SLM to generate executable programs for multi-step operations. AgentCPM-GUI \cite{agentCPM} develops a mobile-use agent through GUI grounding, supervised fine-tuning, and reinforcement fine-tuning, while Ferret-UI Lite \cite{ferret} explores a compact end-to-end agent for interacting with mobile, web, and desktop interfaces. These studies demonstrate the growing capability of compact language models to support complex autonomous interactions on resource-constrained devices. In contrast, PEAM focuses on latency-sensitive communication decision-making, where on-device LLM reasoning must be translated into executable network-configuration actions.

\noindent\textbf{LLM-based Network Configuration}\quad Recent advances in LLMs have also motivated their application to network configuration and mobility management. Lee et al.\cite{roaming_agent} deployed an on-device LLM to interpret WiFi roaming observations and generate corresponding control actions. Nagai et al.\cite{roaming_agent2} further investigated prompt refinement for real-time WLAN handover and evaluated their method using a quantized Llama-3.1-8B model on edge devices. Beyond WiFi roaming, Li et al.\cite{cellular_agent} developed a GPT-2-based framework to model the temporal evolution of reference signal received power, supporting proactive cellular handover. Jeong et al.\cite{cellular_agent2} incorporated visual environmental context through a multi-modal LLM to estimate channel capacity along candidate user paths and assist mobility decisions. These studies show that language models can interpret heterogeneous communication observations and support context-aware network decisions.


\section{Methodology}
As illustrated in Figure~\ref{fig:2}, PEAM consists of four modules: Record Acquisition, Periodic Update, User-Level Local Memory, and Real-Time Execution. We consider personalized mobile network configuration over a sequence of update periods indexed by $t$. In each period, PEAM first collects user-initiated configuration records, then performs offline LLM-based rule generation when the device is idle, updates the local rule and profile memories, and finally executes matched rules online without invoking the LLM.

\subsection{Record Acquisition}
The \textit{Record Acquisition} module runs continuously in the background and monitors user-initiated network-configuration operations. Once an operation is detected, PEAM records the communication context immediately before the operation, the configuration action selected by the user, and the resulting context after the operation. The records collected during the $t$-th update period are represented as
\begin{equation}
    D_t=\left\{d_i^t\right\}_{i=1}^{N_t},
    \qquad
    d_i^t=\left(x_i^{t,-},a_i^t,x_i^{t,+}\right),
    \label{eq:decision_record}
\end{equation}
where $N_t$ denotes the number of collected records; $x_i^{t,-}$ and $x_i^{t,+}$ denote the communication contexts immediately before and after the $i$-th operation, respectively; and $a_i^t$ denotes the configuration action selected by the user. Each context contains information such as time, location, foreground application, device configuration, and network conditions. All records are stored locally until the next periodic update. The collected record set $D_t$, together with the user profile $p_{t-1}$ generated in the previous period, serves as the input to Periodic Updating, which is then used to infer the user's underlying decision intents and generate personalized configuration rules. 


\subsection{Periodic Updating}
\noindent\textbf{Data Clustering}\quad  Records collected within an update period are heterogeneous and may exceed the context capacity of the on-device LLM. PEAM therefore groups $\mathcal{D}_t$ into batches using feature-weighted HDBSCAN\cite{hdbscan}. For two records $d_a$ and $d_b$, the dissimilarity of the $k$-th feature and their overall weighted Gower distance are defined as
\begin{equation}
    \begin{aligned}
        \delta_k(d_a,d_b)
        &=
        \begin{cases}
            \dfrac{|x_a^k-x_b^k|}{R_k}, & k\in\mathcal{F}_{\mathrm{num}},\\[5pt]
            \mathbb{I}(x_a^k\neq x_b^k), & k\in\mathcal{F}_{\mathrm{cat}},
        \end{cases}\\
        \operatorname{dist}_{\mathbf{W}}(d_a,d_b)
        &=
        \frac{\sum_{k=1}^{m}w_k\delta_k(d_a,d_b)}{\sum_{k=1}^{m}w_k}.
    \end{aligned}
\label{eq:weighted_gower}
\end{equation}
where $R_k$ is the range of the numerical feature $k$, $\mathcal{F}_{\mathrm{num}}$ and $\mathcal{F}_{\mathrm{cat}}$ denote the numerical and categorical feature sets, and $w_k$ controls the contribution of each feature. The feature weights $\mathbf W$ and HDBSCAN hyperparameters $\Theta$ are jointly optimized once on labeled training data using the Tree-structured Parzen Estimator\cite{optuna}. Let $\mathbf c^*$ denote the ground-truth intent labels and $\hat{\mathbf c}\equiv\hat{\mathbf c}(\mathbf W,\Theta)$ denote the clustering assignments. The optimization objective is
\begin{equation}
    \begin{aligned}
        Q(\mathbf W,\Theta)
        &=
        \lambda_1\operatorname{ARI}\bigl(\mathbf{c}^{*},\hat{\mathbf{c}}\bigr)
        +
        \lambda_2\operatorname{NMI}\bigl(\mathbf{c}^{*},\hat{\mathbf{c}}\bigr),\\[3pt]
        (\mathbf W^*,\Theta^*)
        &=
        \arg\max_{\mathbf W,\Theta}Q(\mathbf W,\Theta).
    \end{aligned}
\label{eq:clustering_optimization}
\end{equation}
where $\lambda_1+\lambda_2=1$. The optimized parameters are fixed after training and reused during deployment. Applying them to $\mathcal{D}_t$ yields a cluster set $\mathcal{C}_t=\{C_{t,j}\}_{j=1}^{M_t}$, where each cluster forms an independent analysis batch and sparse observations are treated as noise.

\begin{figure}[!htbp]
    \centering
    \includegraphics[width=0.9\linewidth]{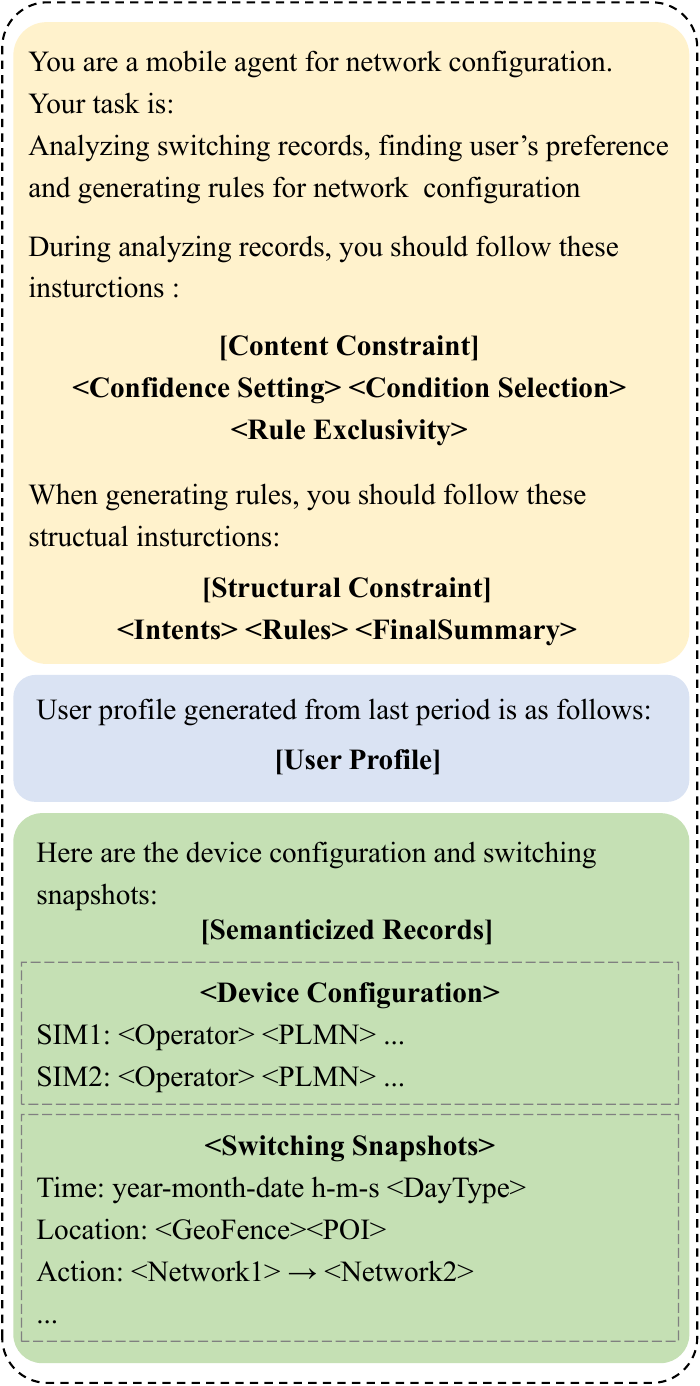}
    \caption{Prompt template for Record Analysis. The yellow boxes contain fixed task and output constraints, the blue box contains the user profile generated in the previous update period, and the green box contains the semantic representations of the current records.}
    \label{fig:3}
\end{figure}

\noindent\textbf{Record Analysis}\quad Although clustering reduces the input length and groups records with similar contexts, raw device logs remain difficult for a compact LLM to interpret directly. PEAM therefore converts each cluster into a concise semantic representation before inference. Stable device attributes, such as SIM information and data-usage settings, are presented once as shared configuration context. Dynamic attributes, such as time, location, and network conditions, are summarized using the mean for numerical features and the mode for categorical features. Each contextual transition is then represented as $K:V^-\rightarrow V^+$, where $K$ denotes an attribute and $V^-$ and $V^+$ denote its values before and after the configuration operation, respectively.

As illustrated in Figure~\ref{fig:3}, the record-analysis prompt combines semantic records with the content constraint $\Gamma_c$, the structural constraint $\Gamma_s$, and the previous user profile $p_{t-1}$. The content constraint guides the model to identify shared behavioral patterns, infer the most plausible user intent and rationale, determine corresponding decision conditions, and verify the compatibility of the resulting configuration action. The structural constraint $\Gamma_s$ enforces a consistent output comprising the inferred intents, confidence scores, rationales, condition-action rules, and a summary. For the $j$-th cluster, the record-analysis process is formulated as
\begin{equation}
    \begin{aligned}
        \left(\hat{\mathcal I}_{t,j},\hat{\mathcal R}_{t,j}\right)
        =
        \operatorname{LLM}\bigl(\Gamma_c,\Gamma_s,\mathbf p_{t-1},\operatorname{Sem}(C_{t,j})\bigr).
    \end{aligned}
\label{eq:record_analysis}
\end{equation}
where $\operatorname{Sem}(\cdot)$ denotes semantic transformation, and $\hat{\mathcal I}_{t,j}$ and $\hat{\mathcal R}_{t,j}$ denote the locally inferred intents and candidate configuration rules, respectively.

\noindent\textbf{Reflection}\quad Clustering cannot guarantee that all records associated with the same underlying intent fall into a single batch. As a result, batch-level outputs may contain incomplete intents, redundant rules, or conflicting actions caused by limited local evidence. PEAM address this problem through cross-batch reflection\cite{reflexion, self_refine}. Guided by a fixed reflection constraint $\Gamma_r$, the LLM identifies intents and rules with shared semantics across batches, merges recurring strategies and complementary conditions to consolidate distributed evidence. It further examines candidate rules with overlapping conditions but incompatible actions, suppressing conclusions supported only by isolated batches. Semantically duplicated rules are also eliminated to prevent fragmented observations from being misinterpreted as independent user preferences. The cross-batch reflection process is formulated as
\begin{equation}
    \begin{aligned}
        (\mathcal I_t,\mathcal R_t)
        =
        \operatorname{LLM}\Big(\Gamma_r,\big\{(\hat{\mathcal I}_{t,j},\hat{\mathcal R}_{t,j})\big\}_{j=1}^{M_t}\Big),
    \end{aligned}
\label{eq:cross_batch_reflection}
\end{equation}
where $\mathcal I_t$ and $\mathcal R_t$ denote the consolidated user intents and configuration rules for the current update period. 

\begin{table*}[t]
    \centering
    \fontsize{9pt}{10pt}\selectfont

    \begin{tabularx}{\textwidth}{
        Y{1.50} |
        *{4}{Y{0.90}} |
        *{2}{Y{0.90}} Y{1.30} |
        Y{0.90} Y{0.90}
    }
        \toprule
        \multicolumn{1}{c|}{\textbf{Method}}
        & \multicolumn{4}{c|}{\textbf{Simulated User Dataset}}
        & \multicolumn{3}{c|}{\textbf{Real User Dataset}}
        & \multicolumn{2}{c}{\textbf{Model Pricing}} \\

        & REA
        & RCR
        & TRR
        & JMS
        & REA
        & RCR
        & TRR
        & Input
        & Output \\
        \midrule

        Expert Rules
        & \underline{94.72}
        & 6.06
        & 8.00
        & 22.15
        & 48.68
        & 2.92
        & 2.56
        & -
        & - \\
        \midrule

        Qwen3.5-Flash
        & 86.58
        & 25.93
        & 56.86
        & 51.90
        & \textbf{74.99}
        & 12.49
        & \textbf{51.35 (19/37)}
        & 0.028
        & 0.278 \\

        Gemini-3-Flash
        & 92.24
        & 54.79
        & 80.19
        & \underline{64.55}
        & \underline{72.51}
        & 15.47
        & \underline{49.02 (25/51)}
        & 0.500
        & 3.000 \\

        GPT-5.2
        & 87.56
        & \underline{57.25}
        & \underline{87.47}
        & 55.72
        & 71.87
        & \underline{20.96}
        & 38.89 (14/36)
        & 1.750
        & 14.00 \\
        \midrule

        PEAM
        & \textbf{99.13}
        & \textbf{60.58}
        & \textbf{92.50}
        & \textbf{81.32}
        & 72.34
        & \textbf{25.86}
        & 38.10 (24/63)
        & -
        & - \\

        \rowcolor{gray!15}
        Improvement
        & +4.41 
        & +3.33 
        & +5.03 
        & +16.77
        & -
        & +4.90 
        & -
        & -
        & - \\
        \bottomrule
    \end{tabularx}

    \caption{End-to-end performance on the simulated-user and real-user datasets, together with API pricing for the cloud-based LLM baselines. REA, RCR, and TRR are reported as percentages. Values in parentheses denote the number of triggered rules over the total number of generated rules. The Improvement row reports the absolute gain over the strongest baseline for each metric. Input and output prices are reported in USD per million tokens. Bold and underlined values denote the best and second-best results, respectively.}
    \label{tab:total_performance}
\end{table*}

\subsection{User-Level Local Memory}
PEAM maintains a user-level local memory consisting of a personalized rule base and a semantic user profile. Each rule $r$ contains a condition--action pair and a confidence score $q_r$. When a rule is retrieved for execution, PEAM asks the user whether to accept the proposed configuration. Acceptance increases the rule confidence, whereas rejection decreases it. Rules whose confidence falls below a predefined threshold are discarded. This feedback mechanism allows PEAM to retain reliable strategies while removing rules that no longer match the user's preferences. Detailed rule states and transition mechanisms are provided in the Appendix.

To summarize user preferences beyond individual condition-specific rules, PEAM periodically constructs a semantic user profile from high-confidence rules. The profile captures both scenario-specific preferences, such as preferred configurations under particular applications, times, locations, and network conditions, and broader behavioral tendencies, such as sensitivity to data usage and preference for connection quality. Let $B_t^{hc}$ denote the high-confidence rules maintained during the $t$-th update period. Guided by a fixed profile-generation constraint $\Gamma_p$, PEAM generates the profile as
\begin{equation}
    p_t = \operatorname{LLM}\left(\Gamma_p, B_t^{hc}\right),
\end{equation}
where $p_t$ denotes the updated user profile. PEAM regenerates the profile from the latest reliable rules and incorporates it into the next offline rule-generation cycle, providing higher-level preference context for subsequent record analysis.

\subsection{Real-Time Execution}
When the communication context changes, PEAM captures the current application, temporal, spatial, device-configuration, and network conditions, and retrieves applicable active rules from the local rule base. Since these rules are generated in advance during periodic updates, the real-time execution path does not invoke the LLM. Instead, PEAM directly compares the current context with the conditions specified by each rule and retrieves the rules whose conditions are satisfied. If multiple rules match, PEAM resolves the conflict using a two-level priority scheme. It first selects the rule with the largest number of matched condition predicates, since this rule provides the more specific description of the current context. If multiple rules contain the same number of matched conditions, the rule with the highest confidence is selected. Finally, PEAM applies the action associated with the selected rule to the device. This retrieval-based design avoids costly real-time model inference and enables millisecond-level online decision-making.

\begin{table*}[t]
    \centering
    \fontsize{9pt}{10pt}\selectfont

    \begin{tabularx}{\textwidth}{
        Y{1.50} |
        *{3}{Y{0.944}} |
        *{3}{Y{0.944}} |
        *{2}{Y{0.944}}
        Y{0.948}
    }
        \toprule
        \textbf{Method}
        & \multicolumn{3}{c|}{\textbf{Record Analysis}}
        & \multicolumn{3}{c|}{\textbf{Reflection}}
        & \multicolumn{3}{c}{\textbf{User Profiling}} \\

        & SS & JMS & IT
        & SS & JMS & IT
        & SS & JMS & IT \\
        \midrule

        Qwen3.5-Flash
        & \underline{0.7451}
        & 74.35
        & \underline{8.73}
        & 0.6577
        & 66.91
        & \textbf{7.34}
        & \underline{0.8726}
        & \underline{87.98}
        & \textbf{2.57} \\

        Gemini-3-Flash
        & 0.6973
        & 79.69
        & \textbf{4.74}
        & 0.6433
        & 64.88
        & 17.31
        & 0.8593
        & 85.54
        & \underline{2.58} \\

        GPT-5.2
        & 0.7418
        & \underline{81.06}
        & 13.80
        & \underline{0.8377}
        & \underline{73.05}
        & \underline{9.76}
        & 0.8582
        & \textbf{89.89}
        & 6.59 \\
        \midrule

        PEAM
        & \textbf{0.8361}
        & \textbf{86.68}
        & 15.64
        & \textbf{0.8955}
        & \textbf{84.12}
        & 13.98
        & \textbf{0.9112}
        & 87.10
        & 5.31 \\

                \rowcolor{gray!15}
        Improvement
        & +0.0910
        & +5.62
        & -
        & +0.0578
        & +11.07
        & -
        & +0.0386
        & -
        & - \\
        \bottomrule
    \end{tabularx}

    \caption{Task-level performance on Record Analysis, Cross-Batch Consolidation, and User Profiling. IT is reported in seconds. The Improvement row reports the absolute gain over the strongest baseline for each metric. Bold and underlined values denote the best and second-best results, respectively.}
    \label{tab:task_performance}
\end{table*}

\begin{table*}[t]
    \centering
    \fontsize{9pt}{10pt}\selectfont

    \begin{tabularx}{\textwidth}{
        Y{1.40} |
        *{3}{Y{0.80}} |
        *{3}{Y{0.80}} |
        *{3}{Y{0.80}} |
        Y{2.40}
    }
        \toprule
        \textbf{Method}
        & \multicolumn{3}{c|}{\textbf{Record Analysis}}
        & \multicolumn{3}{c|}{\textbf{Reflection}}
        & \multicolumn{3}{c|}{\textbf{User Profiling}}
        & \textbf{Real-time Execution}\\

        & SS & JMS & IT
        & SS & JMS & IT
        & SS & JMS & IT 
        & Retrieval Time\\
        \midrule

        PEAM
        & 0.8413
        & 84.06
        & 35.53
        & 0.8563
        & 75.21
        & 32.62
        & 0.8952
        & 84.64
        & 11.87
        & -\\

        PEAM-Q
        & 0.8132
        & 80.80
        & 33.17
        & 0.8608
        & 72.09
        & 59.29
        & 0.7991
        & 64.56
        & 8.950
        & 0.5730 \\

        \bottomrule
    \end{tabularx}

    \caption{Task-level performance of AndesVL-based PEAM under two deployment configurations. PEAM denotes the full-precision model evaluated on an NVIDIA A100 GPU, whereas PEAM-Q denotes the quantized model deployed on a smartphone. Inference time (IT) is reported in seconds, while retrieval time is reported in milliseconds.}
    \label{tab:device_performance}
\end{table*}

\begin{table}[t]
    \centering
    \fontsize{9pt}{10pt}\selectfont

    \begin{tabularx}{\columnwidth}{
        @{}
        l
        Y{1.0}
        Y{1.0}
        Y{1.0}
        Y{1.0}
        @{}
    }
        \toprule
        \textbf{Variant}
        & \textbf{REA}
        & \textbf{RCR}
        & \textbf{TRR}
        & \textbf{JMS} \\
        \midrule

        PEAM
        & \textbf{99.13}
        & 60.58
        & \textbf{92.50}
        & \textbf{81.32} \\

        - w/o Clustering
        & 98.61
        & 49.57
        & 91.79
        & 65.80 \\

        - w/o Reflection
        & 92.44
        & \textbf{67.96}
        & 43.21
        & 70.43 \\

        - w/o User Profiling
        & 97.36
        & 61.31
        & 85.26
        & 75.27 \\

        \bottomrule
    \end{tabularx}

    \caption{Ablation results for the major components of PEAM. REA, RCR, and TRR are reported as percentages.}
    \label{tab:ablation_PEAM_p}
\end{table}

\section{Evaluation}

\subsection{Datasets}
We construct three mutually disjoint datasets for model training and evaluation: a simulated-user dataset, a real-user dataset, and a task-level dataset.

\noindent\textbf{Simulated-User Dataset}\quad We first collected one month of network-configuration records from 51 real users, together with additional records obtained in controlled laboratory settings. These records are manually annotated to identify 20 high-frequency communication intents and their corresponding configuration templates. Based on these annotations, we generate 300 simulated users with diverse spatiotemporal contexts, applications, device configurations, and network conditions, and use DeepSeek-V3.2\cite{deepseek_v32} to annotate the simulated data. Among these users, 200 are used exclusively to optimize the feature weights and HDBSCAN hyperparameters, while the remaining 100 are reserved for end-to-end evaluation.

\noindent\textbf{Real-User Dataset}\quad We further collect approximately two months of network-configuration records from 82 held-out real users. For each user, the records are chronologically divided into two periods. The first period is used to generate personalized rules, while the second is used to evaluate whether these rules correctly predict subsequent user decisions.

\noindent\textbf{Task-Level Dataset}\quad We construct a dialogue dataset covering Record Analysis, Cross-Batch Reflection, and User Profiling. The training set contains 5,000 samples, including 2,000 for Record Analysis, 2,000 for Cross-Batch Reflection, and 1,000 for User Profiling. The test set contains 300 held-out samples, with 100 samples for each task.

\subsection{Baselines and Implementation Details}

\noindent\textbf{Baselines}\quad We compare PEAM with two groups of baselines. The first is an expert-rule method consisting of 11 manually designed rules for common network-switching scenarios. Each rule specifies when a switch should be triggered, while the target action for each user is determined by the configuration most frequently observed in that user’s historical records. The second group includes three general-purpose LLMs: Qwen3.5-Flash \cite{qwen_35}, Gemini-3-Flash \cite{gemini_3_f}, and GPT-5.2 \cite{gpt_52}. In the end-to-end evaluation, these models directly process complete historical records and generate personalized configuration rules under the same output format as PEAM. 

\noindent\textbf{Implementation Details}\quad PEAM uses Qwen3-4B as its default backbone and is fine-tuned with LoRA \cite{lora} on a single NVIDIA A100 GPU. For the deployment analysis, we additionally train an AndesVL-4B-based variant \cite{andesvl}. Its full-precision version is evaluated on the A100 server, while its W4A16-quantized version is deployed on an OPPO Find X9 Pro smartphone equipped with the MediaTek Dimensity 9500. Detailed configurations are provided in the Appendix.

\subsection{Evaluation Metrics}
We evaluate different methods from three perspectives: rule effectiveness, generation quality, and efficiency.

\noindent\textbf{Rule Effectiveness}\quad We use three metrics to evaluate the effectiveness of rules generated from historical records. Rule Execution Accuracy (REA) measures the proportion of correct actions among all triggered rule executions. Rule Coverage Ratio (RCR) measures the proportion of subsequent-period records that trigger at least one generated rule, reflecting the behavioral coverage of the rule set. Triggered Rule Ratio (TRR) measures the proportion of generated rules that are triggered by at least one record in the subsequent evaluation period, indicating whether a method produces excessive inactive rules.

\noindent\textbf{Generation Quality}\quad We use Semantic Similarity (SS) and Judge Model Score (JMS) to measure the discrepancy between generated outputs and their reference labels, where higher scores indicate better generation quality. SS evaluates embedding-level semantic similarity using Qwen3-Embedding-0.6B \cite{qwen3_embed}, while JMS uses DeepSeek-V3.2 to assign a score from 0 to 100 based on their overall consistency. The evaluation criteria for these tasks are summarized in the Appendix.


\noindent\textbf{Efficiency}\quad Inference Time (IT) measures the time required to complete a single inference for a given task. For cloud-based LLM baselines, we also report the input and output prices per million tokens. Since PEAM performs inference locally, it incurs no external API cost.


\subsection{Experiments and Results}
Table 1 reports the end-to-end rule performance on the simulated-user and real-user datasets. On the simulated-user dataset, PEAM achieves the best REA, RCR, TRR, and JMS, demonstrating a strong balance among execution accuracy, behavioral coverage, and rule usability. In contrast, the expert-rule method attains high accuracy only within a small set of predefined scenarios, as reflected by its substantially lower RCR and TRR, while the general-purpose LLMs show less consistent performance across metrics. On the real-user dataset, all methods degrade because of more heterogeneous and less repetitive user behaviors. Nevertheless, PEAM achieves the highest RCR of 25.86\%, corresponding to a 4.90\% improvement over the strongest baseline, while maintaining an REA of 72.34\% that remains comparable to the cloud-based LLMs. Its lower TRR should be interpreted together with the size of the generated rule library: PEAM produces 63 rules, of which 24 are triggered in the subsequent period, nearly matching Gemini-3-Flash despite maintaining a broader candidate set. This lower TRR mainly reflects PEAM’s broader rule exploration, while inactive rules do not affect user experience. Moreover, unlike cloud-based LLMs, PEAM incurs no recurring API cost, yielding a favorable trade-off among accuracy, coverage, and deployment cost.

Table~2 evaluates PEAM on the three offline reasoning tasks: Record Analysis, Cross-Batch Reflection, and User Profiling. PEAM achieves the highest semantic similarity across all three tasks and the best judge scores for Record Analysis and Cross-Batch Reflection, indicating stronger alignment with reference outputs when interpreting communication records and consolidating intermediate results. The clear advantage in Cross-Batch Reflection supports the effectiveness of task-specific fine-tuning for merging distributed evidence and resolving inconsistent rules. For User Profiling, PEAM obtains the best semantic similarity and remains competitive in judge score, suggesting that its generated profiles closely capture reference user preferences. Although PEAM introduces moderate offline inference overhead, these tasks are performed during periodic updates, which occur when the mobile device is idle, and therefore do not affect latency-critical rule execution.

Table 3 compares PEAM under full-precision server inference and W4A16-quantized smartphone deployment. The quantized model largely preserves the quality of Record Analysis and Cross-Batch Reflection, achieving SS values of 0.8132 and 0.8608, respectively. Its reflection similarity is slightly higher than that of the full-precision model, although the corresponding JMS decreases, suggesting that quantization preserves overall semantics but may affect finer-grained generation quality. User Profiling exhibits a more noticeable degradation, indicating that higher-level preference abstraction is more sensitive to quantization than record-level analysis and rule consolidation. Offline inference latency also varies across tasks, showing that quantization does not provide a uniform speedup across the entire pipeline. Most importantly, PEAM-Q completes rule retrieval in only 0.5730 ms on the smartphone. This result directly validates effectiveness of PEAM's separation of offline LLM deliberation and online rule execution in real-time network configuration.


Table 4 evaluates the contribution of each major component in PEAM. Removing clustering reduces RCR from 60.58\% to 49.57\% and decreases JMS from 81.32 to 65.80, while REA and TRR remain relatively high. This is due to that clustering primarily improves evidence organization, behavioral coverage, and generation quality by grouping coherent records before LLM inference. Removing Cross-Batch Reflection increases RCR to 67.96\%, but substantially decreases REA, TRR, and JMS to 92.44\%, 43.21\%, and 70.43, respectively. The increased coverage therefore comes at the cost of many redundant, conflicting, or weakly supported rules, confirming that reflection acts as an essential reliability filter. Removing User Profiling leaves coverage nearly unchanged, but reduces REA, TRR, and JMS to 97.36\%, 85.26\%, and 75.27. This pattern shows that user profiling mainly improves the consistency and personalization of generated rules rather than simply increasing their number. Overall, the three components provide complementary benefits and jointly contribute to PEAM’s performance.

\section{Conclusion}
In this paper, we presented PEAM, an on-device communication agent for personalized mobile network configuration. PEAM decouples LLM-based deliberation from latency-critical execution by distilling historical user decisions into persistent condition–action rules during offline updates and retrieving matched rules online without invoking the LLM. To support compact on-device reasoning over long and heterogeneous histories, PEAM combines batch-wise analysis with cross-batch reflection, while a rule-grounded hybrid memory connects executable rules with semantic user profiles for cross-cycle personalization. Evaluations on task-level, simulated-user, and held-out real-user benchmarks demonstrate competitive performance in rule execution, coverage, triggering, and generation quality, while smartphone deployment of a quantized 4B model confirms its practical on-device feasibility. Overall, PEAM demonstrates that on-device LLM agents can support latency-critical mobile system decisions by shifting complex reasoning offline and executing personalized decisions through efficient local rules.


\bibliography{aaaiRef}

\clearpage
\setlength{\textwidth}{6.5in}
\setlength{\textheight}{9in}
\setlength{\oddsidemargin}{0in}
\setlength{\evensidemargin}{0in}
\setlength{\topmargin}{0in}
\onecolumn
\appendix
\setcounter{secnumdepth}{1}
\setcounter{figure}{0}
\setcounter{table}{0}
\setcounter{equation}{0}
\renewcommand{\thefigure}{A\arabic{figure}}
\renewcommand{\thetable}{A\arabic{table}}
\renewcommand{\theequation}{A\arabic{equation}}
\setlength{\tabcolsep}{4.5pt}
\begin{center}
    {\Large\bfseries Appendix}
\end{center}

\section{Supplementary Experiments}

\subsection{Training Details}
We fine-tune Qwen3-4B and AndesVL-4B through multi-task supervised fine-tuning using LoRA on a single NVIDIA A100 GPU, with a rank of 32, a scaling factor of 32, and a dropout rate of 0.1. We train the model for 5 epochs using AdamW with a learning rate of $5\times10^{-5}$, cosine decay, a warmup ratio of 0.05 and a batch size of 4. We jointly train the model on Record Analysis, Cross-Batch Reflection, and User Profiling. A task-aware sampler enforces a 2:1:1 allocation within each effective batch, corresponding to 2, 1, and 1 samples from the three tasks, respectively. All tasks use equal sample-level loss weights. We further adopt field-aware token weighting to emphasize task-critical outputs. Prompt tokens are excluded from loss computation; within the responses, tokens in the \texttt{rules} field of Record Analysis and the \texttt{result} field of Cross-Batch Reflection are assigned a weight of 10, while all remaining response tokens retain a weight of 1. The weighted token losses are normalized within each sample before being averaged across the batch. This design emphasizes the structured outputs directly used for rule generation and consolidation without uniformly amplifying the loss of an entire task.

\subsection{Baseline Details}
To fully compare PEAM with different methods, we design an expert-rule method consisting of 11 manually designed rules for common network-switching scenarios. Each rule specifies when a switch should be triggered, while the action for each user is determined by the configuration most frequently observed in that user’s historical records. The detailed expert rules are shown in Table~\ref{tab:network-switching-rule-conditions}.

\begin{table}[htbp]
    \centering
    \fontsize{7pt}{7pt}\selectfont
    \renewcommand{\arraystretch}{1.20}

    \begin{tabular*}{1.00\textwidth}{
        @{\extracolsep{\fill}}
        L{0.38\textwidth}
        L{0.58\textwidth}
    }
        \toprule
        \multicolumn{1}{c}{\textbf{Intent}}
        & \multicolumn{1}{c}{\textbf{Condition}} \\
        \midrule

        Optimize cellular network usage for smooth meeting scenarios.
            & \texttt{app\_scene=Meeting}, \texttt{WiFi\_QoE}$<4$, \texttt{cellular\_score}$\geq 80$ \\
        \arrayrulecolor{black!35}
        \specialrule{0.20pt}{1.0pt}{1.0pt}
        \arrayrulecolor{black}

        Optimize cellular network usage for smooth video call scenarios.
            & \texttt{app\_scene=VIDEO\_CALL}, \texttt{WiFi\_QoE}$<4$, \texttt{cellular\_score}$\geq 80$ \\
        \arrayrulecolor{black!35}
        \specialrule{0.20pt}{1.0pt}{1.0pt}
        \arrayrulecolor{black}

        Optimize cellular network usage for smooth voice call scenarios.
            & \texttt{app\_scene=VOICE\_CALL}, \texttt{WiFi\_QoE}$<4$, \texttt{cellular\_score}$\geq 80$ \\
        \arrayrulecolor{black!35}
        \specialrule{0.20pt}{1.0pt}{1.0pt}
        \arrayrulecolor{black}

        Optimize cellular network usage for smooth game-playing scenarios.
            & \texttt{app\_scene=GAME\_PLAYING}, \texttt{WiFi\_QoE}$<4$, \texttt{cellular\_score}$\geq 80$ \\
        \arrayrulecolor{black!35}
        \specialrule{0.20pt}{1.0pt}{1.0pt}
        \arrayrulecolor{black}

        Optimize cellular network usage for smooth game lobby scenarios.
            & \texttt{app\_scene=GAME\_LOBBY}, \texttt{WiFi\_QoE}$<4$, \texttt{cellular\_score}$\geq 80$ \\
        \arrayrulecolor{black!35}
        \specialrule{0.20pt}{1.0pt}{1.0pt}
        \arrayrulecolor{black}

        Save cellular data in live-streaming scenarios.
            & \texttt{app\_scene=LIVE\_STREAM}, \texttt{WiFi\_QoE}$\geq 4$ \\
        \arrayrulecolor{black!35}
        \specialrule{0.20pt}{1.0pt}{1.0pt}
        \arrayrulecolor{black}

        Save cellular data in long-video scenarios.
            & \texttt{app\_scene=LONG\_VIDEO}, \texttt{WiFi\_QoE}$\geq 4$ \\
        \arrayrulecolor{black!35}
        \specialrule{0.20pt}{1.0pt}{1.0pt}
        \arrayrulecolor{black}

        Save cellular data in short-video scenarios.
            & \texttt{app\_scene=SHORT\_VIDEO}, \texttt{WiFi\_QoE}$\geq 4$ \\
        \arrayrulecolor{black!35}
        \specialrule{0.20pt}{1.0pt}{1.0pt}
        \arrayrulecolor{black}

        Save cellular data in download scenarios.
            & \texttt{app\_scene=DOWNLOAD} \\
        \arrayrulecolor{black!35}
        \specialrule{0.20pt}{1.0pt}{1.0pt}
        \arrayrulecolor{black}

        Switch to cellular network when Wi-Fi quality is poor.
            & \texttt{WiFi\_QoE}$<3$, \texttt{cellular\_score}$\geq 80$ \\
        \arrayrulecolor{black!35}
        \specialrule{0.20pt}{1.0pt}{1.0pt}
        \arrayrulecolor{black}

        Switch to Wi-Fi when cellular network quality is poor.
            & \texttt{cellular\_score}$\leq 60$, \texttt{WiFi\_QoE}$\geq 4$ \\

        \bottomrule
    \end{tabular*}

    \caption{Rule Conditions for Network Switching Intents}
    \label{tab:network-switching-rule-conditions}
\end{table}

For the general-purpose models, we design task-specific prompts and retain each model’s default temperature and input/output token limits to reflect its out-of-the-box performance without model-specific tuning. During evaluation, we disable thinking mode for all models to ensure practical and comparable response latency.

\subsection{On-Device Deployment}
\begin{figure}[htbp]
    \centering
    \includegraphics[width=0.95\linewidth]{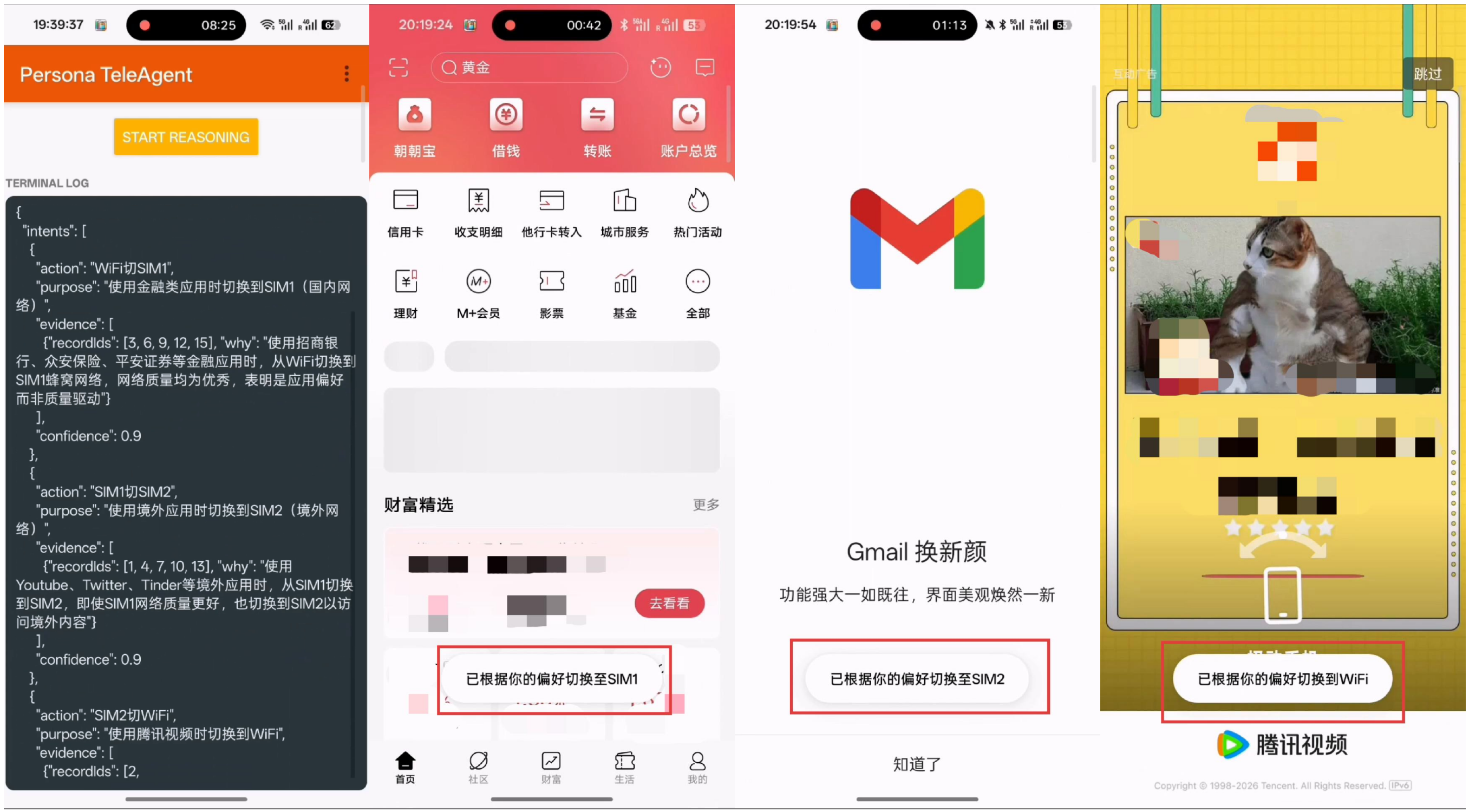}
    \caption{On-device deployment of PEAM.}
    \label{fig:demo}
\end{figure}

To validate on-device deployment, we ran a PEAM variant based on quantized AndesVL on an OPPO Find X9 Pro (MediaTek Dimensity 9500 processor, 16 GB of RAM, Android 16). The model adopts mixed-precision quantization: the backbone linear layers use W4A16, with 4-bit asymmetric weights and 16-bit symmetric activations; the KV cache uses INT8; RMSNorm, the final normalization layer, and logits use INT16; and the embedding and softmax layers retain floating-point precision. The left side of Figure~\ref{fig:demo} shows the logs produced when PEAM analyzes a user’s network-switching records and generates personalized strategies. The right side illustrates how PEAM automatically applies a previously generated strategy when the same context recurs, such as when the user launches a particular application. The overall performance results demonstrate the feasibility of on-device inference and real-time rule execution on consumer-grade hardware.

We further evaluated rule-execution efficiency on a device running PEAM. During the experiment, participants first performed several prescribed network switches, allowing PEAM to generate corresponding rules. After rule generation, they triggered the same scenarios again so that PEAM could retrieve and execute the matched rules. As shown in Table~\ref{tab:retrieval_execution_time}, the average rule-retrieval latency during real-time execution was 0.573 ms. Applying the selected rule required an average of 614.2 ms for switching between cellular networks and 93.57 ms for switching between cellular and Wi-Fi networks. Overall, PEAM completed rule retrieval and application within one second in all tested scenarios, satisfying users’ responsiveness requirements.

\begin{table}[htbp]
    \centering
    \fontsize{9pt}{11pt}\selectfont

    \begin{tabularx}{\textwidth}{
        >{\centering\arraybackslash}X |
        >{\centering\arraybackslash}X |
        >{\centering\arraybackslash}X
        >{\centering\arraybackslash}X
    }
        \toprule
        \textbf{Metric}
        & \textbf{Retrieval Time}
        & \multicolumn{2}{c}{\textbf{Execution Time}} \\
        \cmidrule(lr){3-4}

        &
        & \textbf{Cellular-to-Cellular}
        & \textbf{Cellular-to-Wi-Fi} \\
        \midrule

        Time (ms)
        & 0.573
        & 614.2
        & 93.57 \\
        \bottomrule
    \end{tabularx}

    \caption{Runtime of strategy retrieval and execution.
    All times are reported in milliseconds.}
    \label{tab:retrieval_execution_time}
\end{table}

\subsection{Real Examples}
We continuously monitored several users and collected at least one month of data from their mobile devices. In subsequent interviews, we asked the users to explain their motivations for performing network switches. We then compared the interview findings with the results produced by PEAM. The detailed comparison is presented in Tables~\ref{tab:user001-comparison}--\ref{tab:user006-comparison}.
\begin{table}[htbp]
    \centering
    \fontsize{7pt}{7pt}\selectfont
    \renewcommand{\arraystretch}{1.20}
    \begin{tabular*}{1.00\textwidth}{@{\extracolsep{\fill}} L{0.55\textwidth} C{0.20\textwidth} C{0.20\textwidth}}
        \toprule
        \multicolumn{1}{c}{\textbf{Intent}} & \textbf{Objective Evaluation} & \textbf{Subjective Evaluation} \\
        \midrule
        \multicolumn{3}{l}{\textit{Interview Results}} \\
        \midrule
        When using Douyin at two regular home locations, prefer connecting to the home Wi-Fi network; switch to cellular data when Wi-Fi is unavailable. & --- & --- \\
        \arrayrulecolor{black!35}\specialrule{0.20pt}{1.0pt}{1.0pt}\arrayrulecolor{black}
        When using Nike Run Club at home, prefer connecting to Wi-Fi. & --- & --- \\
        \arrayrulecolor{black!35}\specialrule{0.20pt}{1.0pt}{1.0pt}\arrayrulecolor{black}
        At home, prefer connecting to the home Wi-Fi network regardless of the app being used. & --- & --- \\
        \midrule
        \multicolumn{3}{l}{\textit{PEAM Analysis Results}} \\
        \midrule
        Switch to the cellular network when using Douyin or watching short videos at specific locations such as POI-1 and POI-2 to ensure a smooth viewing experience. & Fully Compliant (55) & Fully Compliant \\
        \arrayrulecolor{black!35}\specialrule{0.20pt}{1.0pt}{1.0pt}\arrayrulecolor{black}
        Prefer connecting to a specific Wi-Fi network, SSID-1, at specific locations such as POI-1 and POI-3 to save cellular data or obtain a better network experience. & Fully Compliant (64) & Fully Compliant \\
        \bottomrule
    \end{tabular*}
    \caption{Comparison Between User Interview Results and PEAM Analysis Results for User 001}
    \label{tab:user001-comparison}
\end{table}

\begin{table}[htbp]
    \centering
    \fontsize{7pt}{7pt}\selectfont
    \renewcommand{\arraystretch}{1.20}
    \begin{tabular*}{1.00\textwidth}{@{\extracolsep{\fill}} L{0.55\textwidth} C{0.20\textwidth} C{0.20\textwidth}}
        \toprule
        \multicolumn{1}{c}{\textbf{Intent}} & \textbf{Objective Evaluation} & \textbf{Subjective Evaluation} \\
        \midrule
        \multicolumn{3}{l}{\textit{Interview Results}} \\
        \midrule
        Use the China Mobile app for faster login; switch to the corresponding carrier SIM when opening the app, and switch back to Wi-Fi after login. & --- & --- \\
        \arrayrulecolor{black!35}\specialrule{0.20pt}{1.0pt}{1.0pt}\arrayrulecolor{black}
        Connect to Wi-Fi when downloading apps from the app store to save cellular data. & --- & --- \\
        \arrayrulecolor{black!35}\specialrule{0.20pt}{1.0pt}{1.0pt}\arrayrulecolor{black}
        Switch to the SIM 1 cellular network before using Alipay to ensure network stability. & --- & --- \\
        \arrayrulecolor{black!35}\specialrule{0.20pt}{1.0pt}{1.0pt}\arrayrulecolor{black}
        Connect to the commonly used Wi-Fi network when entering a regular location. & --- & --- \\
        \arrayrulecolor{black!35}\specialrule{0.20pt}{1.0pt}{1.0pt}\arrayrulecolor{black}
        Switch back to a more stable cellular network when Wi-Fi quality is poor. & --- & --- \\
        \arrayrulecolor{black!35}\specialrule{0.20pt}{1.0pt}{1.0pt}\arrayrulecolor{black}
        Prefer using the cellular network when using the TeamTalk app at home. & --- & --- \\
        \midrule
        \multicolumn{3}{l}{\textit{PEAM Analysis Results}} \\
        \midrule
        Prefer connecting to SSID-1 at home, POI-1, on holiday afternoons. & Fully Compliant (7) & Partially Compliant \\
        \arrayrulecolor{black!35}\specialrule{0.20pt}{1.0pt}{1.0pt}\arrayrulecolor{black}
        Prefer connecting to the company Wi-Fi network, SSID-2, at the office location POI-2 on weekday afternoons. & Fully Compliant (13) & Partially Compliant \\
        \arrayrulecolor{black!35}\specialrule{0.20pt}{1.0pt}{1.0pt}\arrayrulecolor{black}
        Switch to the cellular network when the Wi-Fi user experience score is poor or relatively low to obtain a stable network connection. & Fully Compliant (31) & Fully Compliant \\
        \arrayrulecolor{black!35}\specialrule{0.20pt}{1.0pt}{1.0pt}\arrayrulecolor{black}
        Connect to SSID-1 at POI-3 on holiday mornings. & Non-Compliant (0) & Partially Compliant \\
        \arrayrulecolor{black!35}\specialrule{0.20pt}{1.0pt}{1.0pt}\arrayrulecolor{black}
        Connect to SSID-1 when using Sina Weibo on holiday afternoons. & Partially Compliant (1) & Not Mentioned \\
        \bottomrule
    \end{tabular*}
    \caption{Comparison Between User Interview Results and PEAM Analysis Results for User 002}
    \label{tab:user002-comparison}
\end{table}

\begin{table}[htbp]
    \centering
    \fontsize{7pt}{7pt}\selectfont
    \renewcommand{\arraystretch}{1.20}
    \begin{tabular*}{1.00\textwidth}{@{\extracolsep{\fill}} L{0.55\textwidth} C{0.20\textwidth} C{0.20\textwidth}}
        \toprule
        \multicolumn{1}{c}{\textbf{Intent}} & \textbf{Objective Evaluation} & \textbf{Subjective Evaluation} \\
        \midrule
        \multicolumn{3}{l}{\textit{Interview Results}} \\
        \midrule
        When using Xiaohongshu in the first regular area at home, prefer manually connecting to the home Wi-Fi network. & --- & --- \\
        \arrayrulecolor{black!35}\specialrule{0.20pt}{1.0pt}{1.0pt}\arrayrulecolor{black}
        When using Bilibili in the first regular area at the office, prefer manually connecting to an available Wi-Fi network; switch to cellular when the network quality is poor. & --- & --- \\
        \midrule
        \multicolumn{3}{l}{\textit{PEAM Analysis Results}} \\
        \midrule
        Connect to the cellular network when entering the company location. & Partially Compliant (2) & Not Mentioned \\
        \arrayrulecolor{black!35}\specialrule{0.20pt}{1.0pt}{1.0pt}\arrayrulecolor{black}
        Habitually switch to Wi-Fi when using Xiaohongshu at the company location POI-1. & Partially Compliant (1) & Not Mentioned \\
        \arrayrulecolor{black!35}\specialrule{0.20pt}{1.0pt}{1.0pt}\arrayrulecolor{black}
        Habitually switch to SIM 1 when using system tools such as Performance Monitor at POI-2. & Partially Compliant (1) & Not Mentioned \\
        \bottomrule
    \end{tabular*}
    \caption{Comparison Between User Interview Results and PEAM Analysis Results for User 003}
    \label{tab:user003-comparison}
\end{table}

\begin{table}[htbp]
    \centering
    \fontsize{7pt}{7pt}\selectfont
    \renewcommand{\arraystretch}{1.20}
    \begin{tabular*}{1.00\textwidth}{@{\extracolsep{\fill}} L{0.55\textwidth} C{0.20\textwidth} C{0.20\textwidth}}
        \toprule
        \multicolumn{1}{c}{\textbf{Intent}} & \textbf{Objective Evaluation} & \textbf{Subjective Evaluation} \\
        \midrule
        \multicolumn{3}{l}{\textit{Interview Results}} \\
        \midrule
        Prefer using SIM 1 for Douyin short videos, possibly because SIM 1 has sufficient cellular data. & --- & --- \\
        \arrayrulecolor{black!35}\specialrule{0.20pt}{1.0pt}{1.0pt}\arrayrulecolor{black}
        Switch to SIM 2 for Douyin and TeamTalk when the SIM 1 network quality is poor. & --- & --- \\
        \arrayrulecolor{black!35}\specialrule{0.20pt}{1.0pt}{1.0pt}\arrayrulecolor{black}
        Switch to the cellular network when Wi-Fi quality is poor while using apps such as Youdao Dictionary and WeChat. & --- & --- \\
        \arrayrulecolor{black!35}\specialrule{0.20pt}{1.0pt}{1.0pt}\arrayrulecolor{black}
        Proactively switch to the cellular network before leaving a regular location, likely to avoid lag caused by weak signals near the edge of Wi-Fi coverage. & --- & --- \\
        \arrayrulecolor{black!35}\specialrule{0.20pt}{1.0pt}{1.0pt}\arrayrulecolor{black}
        Connect to another phone's hotspot for internet access during phone data backup or data migration. & --- & --- \\
        \arrayrulecolor{black!35}\specialrule{0.20pt}{1.0pt}{1.0pt}\arrayrulecolor{black}
        Prefer using Wi-Fi when updating or downloading apps to save cellular data. & --- & --- \\
        \midrule
        \multicolumn{3}{l}{\textit{PEAM Analysis Results}} \\
        \midrule
        Habitually switch to the cellular network when using Douyin at night. & Mostly Non-Compliant (0) & Partially Compliant \\
        \arrayrulecolor{black!35}\specialrule{0.20pt}{1.0pt}{1.0pt}\arrayrulecolor{black}
        Prefer using the cellular network when using WeChat on weekday mornings. & Fully Compliant (11) & Not Mentioned \\
        \arrayrulecolor{black!35}\specialrule{0.20pt}{1.0pt}{1.0pt}\arrayrulecolor{black}
        Prefer connecting to SSID-1 at home, POI-1. & Fully Compliant (9) & Not Mentioned \\
        \arrayrulecolor{black!35}\specialrule{0.20pt}{1.0pt}{1.0pt}\arrayrulecolor{black}
        Prefer using the cellular network when using TeamTalk on weekday afternoons. & Fully Compliant (10) & Partially Compliant \\
        \bottomrule
    \end{tabular*}
    \caption{Comparison Between User Interview Results and PEAM Analysis Results for User 004}
    \label{tab:user004-comparison}
\end{table}

\begin{table}[htbp]
    \centering
    \fontsize{7pt}{7pt}\selectfont
    \renewcommand{\arraystretch}{1.20}
    \begin{tabular*}{1.00\textwidth}{@{\extracolsep{\fill}} L{0.55\textwidth} C{0.20\textwidth} C{0.20\textwidth}}
        \toprule
        \multicolumn{1}{c}{\textbf{Intent}} & \textbf{Objective Evaluation} & \textbf{Subjective Evaluation} \\
        \midrule
        \multicolumn{3}{l}{\textit{Interview Results}} \\
        \midrule
        When cellular data becomes stuck, prefer switching to a Wi-Fi network with better signal quality to continue using apps. & --- & --- \\
        \arrayrulecolor{black!35}\specialrule{0.20pt}{1.0pt}{1.0pt}\arrayrulecolor{black}
        Manually connect to the in-building Wi-Fi network at POI-1. & --- & --- \\
        \arrayrulecolor{black!35}\specialrule{0.20pt}{1.0pt}{1.0pt}\arrayrulecolor{black}
        Prefer using the company Wi-Fi network when using TeamTalk at the office. & --- & --- \\
        \arrayrulecolor{black!35}\specialrule{0.20pt}{1.0pt}{1.0pt}\arrayrulecolor{black}
        Use cellular data to access Tencent News when Wi-Fi is unavailable. & --- & --- \\
        \arrayrulecolor{black!35}\specialrule{0.20pt}{1.0pt}{1.0pt}\arrayrulecolor{black}
        When playing DNF Mobile, switch to cellular data if the connected Wi-Fi network has poor quality. & --- & --- \\
        \midrule
        \multicolumn{3}{l}{\textit{PEAM Analysis Results}} \\
        \midrule
        Switch to the SIM 1 cellular network when the Wi-Fi user experience score is poor or relatively low to obtain a stable network connection. & Fully Compliant (4) & Partially Compliant \\
        \bottomrule
    \end{tabular*}
    \caption{Comparison Between User Interview Results and PEAM Analysis Results for User 005}
    \label{tab:user005-comparison}
\end{table}

\begin{table}[htbp]
    \centering
    \fontsize{7pt}{7pt}\selectfont
    \renewcommand{\arraystretch}{1.20}
    \begin{tabular*}{1.00\textwidth}{@{\extracolsep{\fill}} L{0.55\textwidth} C{0.20\textwidth} C{0.20\textwidth}}
        \toprule
        \multicolumn{1}{c}{\textbf{Intent}} & \textbf{Objective Evaluation} & \textbf{Subjective Evaluation} \\
        \midrule
        \multicolumn{3}{l}{\textit{Interview Results}} \\
        \midrule
        When using Douyin, switch to cellular data if the connected Wi-Fi network becomes stuck. & --- & --- \\
        \midrule
        \multicolumn{3}{l}{\textit{PEAM Analysis Results}} \\
        \midrule
        Switch to the SIM 2 cellular network when the Wi-Fi user experience score is poor or relatively low to obtain a stable network connection. & Fully Compliant (4) & Partially Compliant \\
        \arrayrulecolor{black!35}\specialrule{0.20pt}{1.0pt}{1.0pt}\arrayrulecolor{black}
        Switch to the cellular network when watching Douyin live streams on holiday mornings to ensure smooth viewing and avoid lag. & Partially Compliant (2) & Partially Compliant \\
        \arrayrulecolor{black!35}\specialrule{0.20pt}{1.0pt}{1.0pt}\arrayrulecolor{black}
        Connect to SSID-1 at POI-1. & Fully Compliant (9) & Not Mentioned \\
        \bottomrule
    \end{tabular*}
    \caption{Comparison Between User Interview Results and PEAM Analysis Results for User 006}
    \label{tab:user006-comparison}
\end{table}

\subsection{Generation Quality Evaluation}
For Judge Model Score (JMS), DeepSeek-V3.2 compares each generated output with its reference answer and assigns an overall score from 0 to 100. For Record Analysis and Cross-Batch Reflection, the evaluation considers decision-structure completeness, intent-identification accuracy, and agreement of the generated network decisions with the reference. For User Profiling, it considers semantic consistency and information completeness, factual accuracy of network preferences and scenario descriptions, and naturalness and fluency of expression.

\clearpage
\section{Dataset}

\subsection{Real-World Record Acquisition}
To characterize the communication contexts in which users perform network switches, we recruited more than 1,00 participants and, with their consent, installed a data-collection application on their smartphones. To comprehensively describe each communication context, we collected multiple types of data covering spatiotemporal information, foreground applications, network access states, cellular network quality, and Wi-Fi network quality, as detailed in Table~\ref{tab:contextual-features}. It is worth noting that all collected records are conducted with the participants’ explicit consent and authorization. Reocrds are strictly encrypted both at rest and in transit to safeguard user privacy and prevent unauthorized disclosure.

\begin{table}[htbp]
    \centering
    \fontsize{7pt}{7pt}\selectfont
    \renewcommand{\arraystretch}{1.20}

    \begin{tabular*}{1.00\textwidth}{
        @{\extracolsep{\fill}}
        L{0.25\textwidth}
        L{0.70\textwidth}
    }
        \toprule
        \multicolumn{1}{c}{\textbf{Category}}
        & \multicolumn{1}{c}{\textbf{Features}} \\
        \midrule

        Spatiotemporal Context
            & Timestamp, POI, geofence \\
        \arrayrulecolor{black!35}
        \specialrule{0.20pt}{1.0pt}{1.0pt}
        \arrayrulecolor{black}

        Foreground Application
            & App package name, app scene, app foreground dwell time, app-scene foreground dwell time \\
        \arrayrulecolor{black!35}
        \specialrule{0.20pt}{1.0pt}{1.0pt}
        \arrayrulecolor{black}

        Network Access Status
            & Network access type, data SIM slot, PLMN, roaming status \\
        \arrayrulecolor{black!35}
        \specialrule{0.20pt}{1.0pt}{1.0pt}
        \arrayrulecolor{black}

        Cellular Network
            & Cellular network type, RSRP, RSRQ, SINR, cellular signal level, network quality score \\
        \arrayrulecolor{black!35}
        \specialrule{0.20pt}{1.0pt}{1.0pt}
        \arrayrulecolor{black}

        Wi-Fi Network
            & SSID, BSSID, frequency band, bandwidth, RSSI, QoE score, Wi-Fi signal level \\
        \arrayrulecolor{black!35}
        \specialrule{0.20pt}{1.0pt}{1.0pt}
        \arrayrulecolor{black}

        Others
            & Traffic configuration, VPN configuration, battery status \\

        \bottomrule
    \end{tabular*}

    \caption{Contextual Features Used for Network Intent Modeling}
    \label{tab:contextual-features}
\end{table}

The data-collection application operates similarly to the Record Acquisition module in PEAM. When a user initiates a network switch, the application records the buffered data from the preceding $\delta\left(T\right)$ seconds at one-second granularity and continues collecting data for $\delta\left(T\right)$ seconds after the switch. The two temporal windows are combined into a snapshot of a single switching event and stored locally on the device. After operating for a predefined collection period, the application automatically encrypts and uploads the accumulated snapshots, which constitute the user’s network-switching records for that period.

\subsection{Data Preprocessing}
\begin{figure}[htbp]
    \centering
    \includegraphics[width=\linewidth,height=0.94\textheight,keepaspectratio]{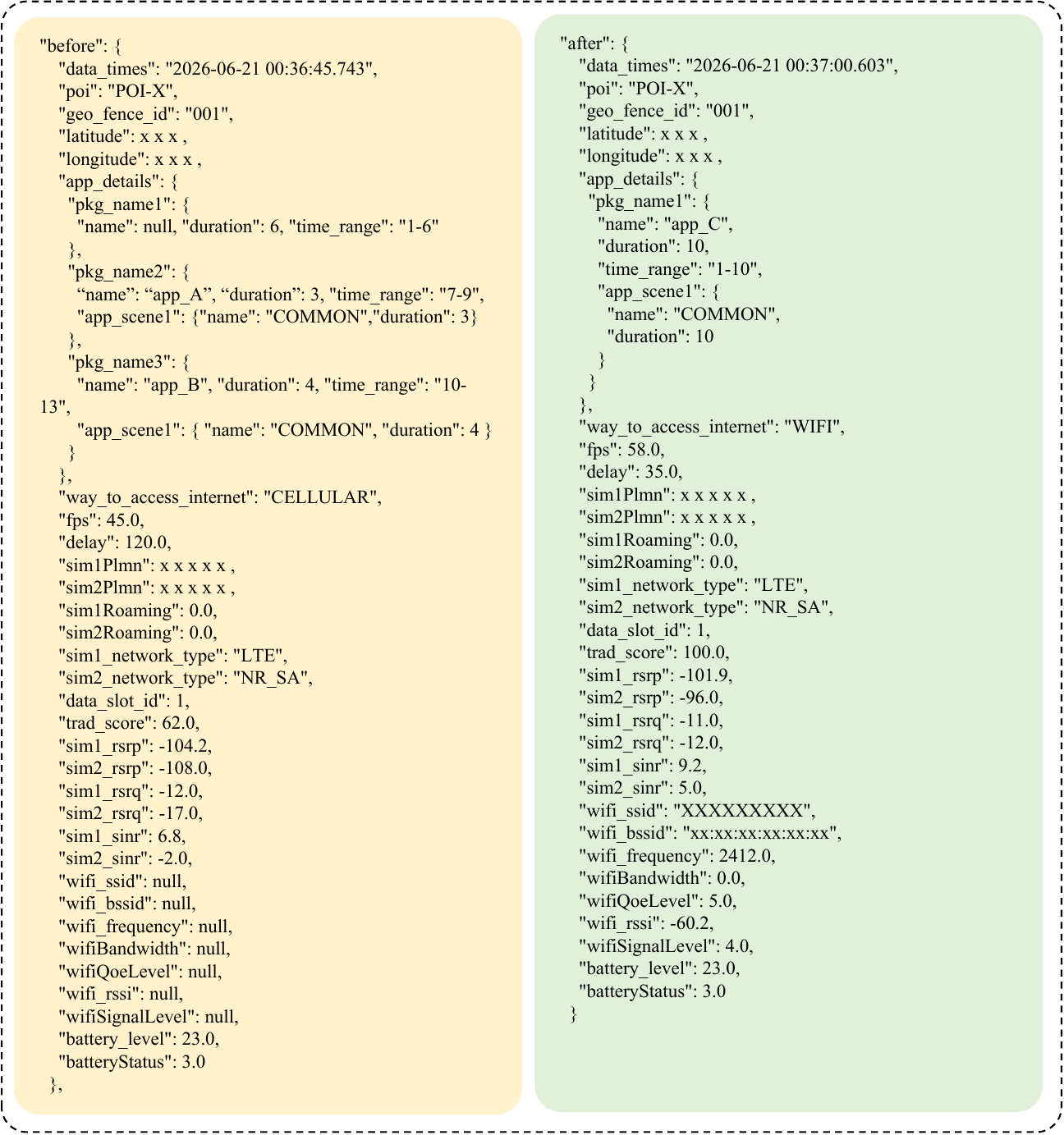}
    \caption{An example of the resulting standardized data representation}
    \label{fig:sample-data}
\end{figure}

Because the communication context is recorded at one-second granularity, even a single switching record can result in a lengthy textual representation. Meanwhile, each per-second snapshot provides only an instantaneous view, making it difficult for an LLM to directly identify temporal changes. We therefore statistically summarize the raw data before presenting it to the model. We categorize the collected fields into three types: stable fields, continuously varying fields (e.g., RSRP and RSRQ), and discretely varying fields (e.g., SSID). Stable fields mainly include temporal and location-related information, which typically exhibits no significant change around a network switch. We therefore use the value observed at the switching moment as the representative value for the entire snapshot. For continuously varying fields, we compute the mean separately over the pre-switch and post-switch windows. For discretely varying fields, we use the mode within each window. Because the foreground application provides the most direct indication of user–device interaction, we preserve the complete foreground-application trace throughout the observation window rather than aggregating it. Figure~\ref{fig:sample-data} illustrates an example of the resulting standardized data representation.

The statistically summarized data still cannot be directly provided to the model, as compact LLMs often struggle to interpret raw numerical communication metrics. We therefore apply a semantic transformation to make the underlying context and its changes more explicit. For each record, the transformation program first reads the pre-switch and post-switch values and then processes each field individually. For the temporal field, it extracts the switching time and determines whether the date is a working day and which period of the day the event belongs to. Location is represented by combining geofence and point-of-interest (POI) information. For foreground applications, the program maps package names to human-readable application names and lists the application changes over time line by line. Numerical network-quality metrics are converted into qualitative descriptions, allowing the model to interpret changes in network conditions more intuitively. Following this procedure, each field transition is expressed in the form $A \rightarrow B$, producing a semantic representation of the complete switching process. An example of the standardized semantic representation is shown below.

\begin{figure}[htbp]
    \centering
    \includegraphics[width=\linewidth,height=0.94\textheight,keepaspectratio]{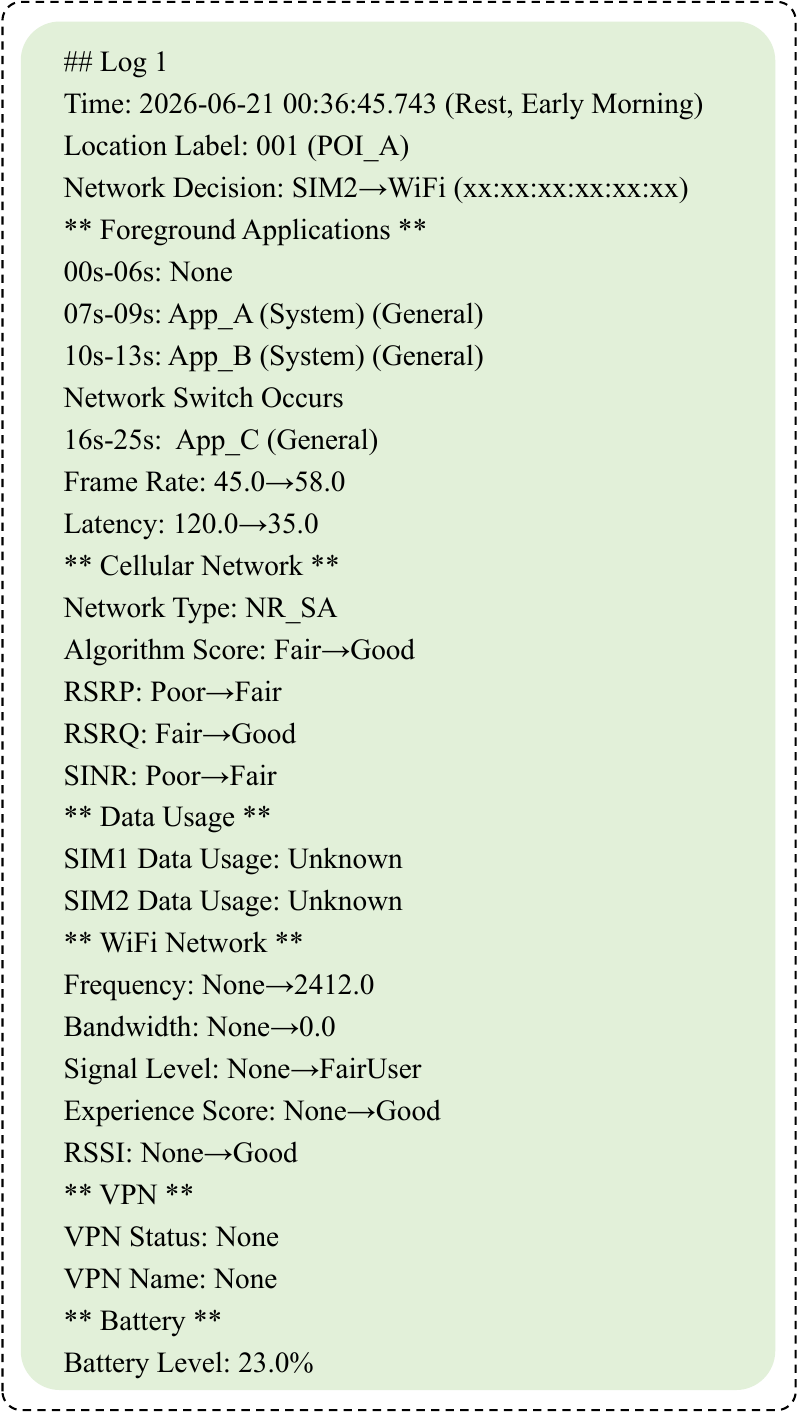}
    \caption{An example of the standardized semantic representation.}
    \label{fig:sample-log}
\end{figure}

\subsection{Simulation}
We construct simulated-user records from templates derived from real-world network-switching samples, together with user-specific contexts and annotated intent rules. The simulated users vary in their spatiotemporal contexts, applications, device configurations, and network conditions. Each record represents a before--after network-state transition associated with a condition--action rule, retaining the general structure of the collected records. The resulting data provide explicit intent and rule annotations for optimizing clustering on the simulated training users and evaluating personalized rule generation on the disjoint simulated test users.

\clearpage
\section{Clustering Algorithm}
We employ a Tree-structured Parzen Estimator (TPE)-based hyperparameter optimization strategy on the simulated training users to jointly optimize the feature weights and HDBSCAN parameters. The clustering objective combines the Adjusted Rand Index (ARI) and Normalized Mutual Information (NMI) with weights of 0.7 and 0.3, respectively. TPE iteratively searches for the parameter configuration that maximizes the average clustering score across users.

We then apply the optimized parameters to the simulated test users to evaluate clustering performance. The results show that the resulting clustering configuration can effectively separate records associated with different latent user intents, thereby improving input coherence and reducing noise in the data provided to the model.

\begin{figure}[htbp]
    \centering
    \includegraphics[width=1.0\linewidth]{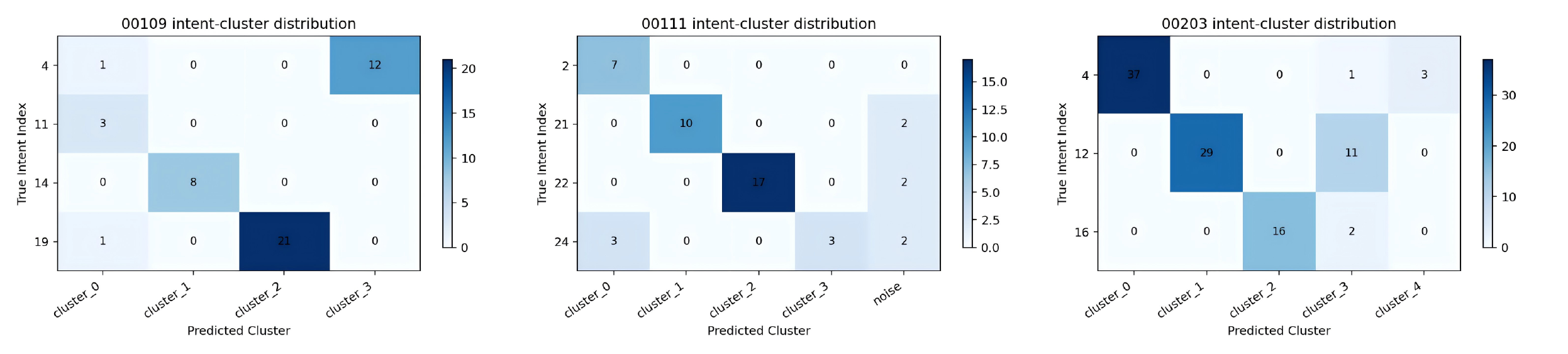}
    \caption{Clustering results of optimized parameters on simulated users.}
    \label{fig:clustering-results}
\end{figure}

\FloatBarrier
\section{User-Level Local Memory Management}
Each rule $r$ is represented by a condition--action pair, an LLM-assigned confidence score $q_r$, and a state $s_r\in\{\textit{silent},\textit{active},\textit{invalid}\}$. Although the LLM provides an initial confidence score $q_r^{0}$ during rule generation, every newly generated rule enters the rule base in the silent state. A silent rule can participate in condition matching but cannot recommend or execute its action. Instead, when its condition is satisfied, PEAM passively observes the network configuration selected by the user without presenting a recommendation. If the user's choice agrees with the rule action, the observation is regarded as positive evidence; otherwise, it is treated as negative evidence. A new rule is activated only after two consecutive positive observations, which prevents rules supported by incidental or noisy records from being immediately applied.

PEAM updates rule confidence using fixed positive and negative steps. Let $y_r\in\{0,1\}$ denote the feedback associated with a matched rule, where $y_r=1$ indicates positive evidence and $y_r=0$ indicates negative evidence. The confidence update is defined as
\begin{equation}
    q_r \leftarrow
    \begin{cases}
        q_r+\Delta^{+}, & y_r=1,\\
        q_r-\Delta^{-}, & y_r=0,
    \end{cases}
\end{equation}
where $\Delta^{+}$ and $\Delta^{-}$ denote the confidence increase and decrease steps, respectively. For a silent rule, $y_r$ is determined by whether the user independently selects the same network configuration as the rule action. Any inconsistent observation resets the consecutive-match counter. For an active rule, PEAM presents the retrieved configuration to the user and asks whether it should be accepted. Acceptance is treated as positive feedback, whereas rejection or the absence of feedback is treated as negative feedback. The configuration is executed only after explicit acceptance.

Active rules are eligible for real-time recommendation and execution. When the confidence of an active rule falls below the activation threshold $\tau_a$, the rule returns to the silent state and is subsequently evaluated through passive observations. Unlike newly generated rules, a previously active rule can be reactivated once its confidence recovers to $\tau_a$, without requiring two additional consecutive matches. If the confidence falls below the invalidation threshold $\tau_i$, the rule becomes invalid and is permanently excluded from subsequent matching, execution, and user-profile construction. Thus, active and silent states are mutually reversible, whereas the invalid state is irreversible.

When multiple active rules match the current context, PEAM first selects the rule whose condition contains the largest number of matched fields. If several rules have the same number of matched fields, the rule with the highest confidence is selected. Only the selected rule receives the current user feedback and updates its confidence. The user profile is periodically reconstructed using active rules only, ensuring that it is grounded in validated and currently reliable configuration preferences.


\end{document}